\documentclass[letterpaper, 10 pt, conference]{ieeeconf}  % Comment this line out if you need a4paper

\IEEEoverridecommandlockouts                              % This command is only needed if 
\usepackage{amsmath}
\usepackage{amssymb}
\usepackage{booktabs}
\usepackage{graphicx}
\usepackage{multirow}
\usepackage{xcolor}
\usepackage{color,colortbl}
\usepackage{subcaption}
\usepackage{bm}
\usepackage{soul}
\usepackage{hyperref}
\hypersetup{
    colorlinks=true,
    linkcolor=blue,
    urlcolor=magenta
    }

\definecolor{beaublue}{rgb}{0.74, 0.83, 0.9}

\title{\LARGE \bf
Action Segmentation Using 2D Skeleton Heatmaps and Multi-Modality Fusion
}

\author{Syed Waleed Hyder~~~~~Muhammad Usama~~~~~Anas Zafar~~~~~Muhammad Naufil~~~~~Fawad Javed Fateh\\Andrey Konin~~~~~M. Zeeshan Zia~~~~~Quoc-Huy Tran
	\thanks{All authors are with Retrocausal, Inc., Redmond, WA 98052, USA.\newline
	Website: \url{www.retrocausal.ai}\newline
	Email: {\tt\small \{waleed,shaik,anas,naufil,fawad,andrey,\newline
	zeeshan,huy\}@retrocausal.ai}}
}

\begin{document}

\maketitle
\thispagestyle{empty}
\pagestyle{empty}

%%%%%%%%%%%%%%%%%%%%%%%%%%%%%%%%%%%%%%%%%%%%%%%%%%%%%%%%%%%%%%%%%%%%%%%%%%%%%%%%
\begin{abstract}
This paper presents a 2D skeleton-based action segmentation method with applications in fine-grained human activity recognition. In contrast with state-of-the-art methods which directly take sequences of 3D skeleton coordinates as inputs and apply Graph Convolutional Networks (GCNs) for spatiotemporal feature learning, our main idea is to use sequences of 2D skeleton heatmaps as inputs and employ Temporal Convolutional Networks (TCNs) to extract spatiotemporal features. Despite lacking 3D information, our approach yields comparable/superior performances and better robustness against missing keypoints than previous methods on action segmentation datasets. Moreover, we improve the performances further by using both 2D skeleton heatmaps and RGB videos as inputs. To our best knowledge, this is the first work to utilize 2D skeleton heatmap inputs and the first work to explore 2D skeleton+RGB fusion for action segmentation.
\end{abstract}

%%%%%%%%%%%%%%%%%%%%%%%%%%%%%%%%%%%%%%%%%%%%%%%%%%%%%%%%%%%%%%%%%%%%%%%%%%%%%%%%
\section{Introduction}
\label{sec:introduction}

With the arrival of advanced deep networks and large-scale datasets, action recognition~\cite{tran2015learning,feichtenhofer2019slowfast}, which aims to classify a trimmed video into a single action label, has achieved considerable progress and maturity. However, action segmentation~\cite{lea2017temporal,filtjens2022skeleton,ding2022temporal}, which seeks to locate and classify action segments of an untrimmed video into action labels, remains a challenging problem. Action segmentation underpins a number of robotics and computer vision applications. Notable examples include human-robot interaction~\cite{rea2019human,khan2022timestamp,yang2023is} (i.e., recognizing human actions to facilitate interactions between humans and robots), ergonomics studies~\cite{parsa2020spatio,parsa2021multi} (i.e., extracting action segments in videos for ergonomics analyses), and visual analytics~\cite{ji2020motion,ji2022computer} (i.e., conducting time and motion studies on video recordings).

\begin{figure}[t]
    \centering
	\includegraphics[width=0.9\linewidth, trim = 0mm 20mm 145mm 0mm, clip]{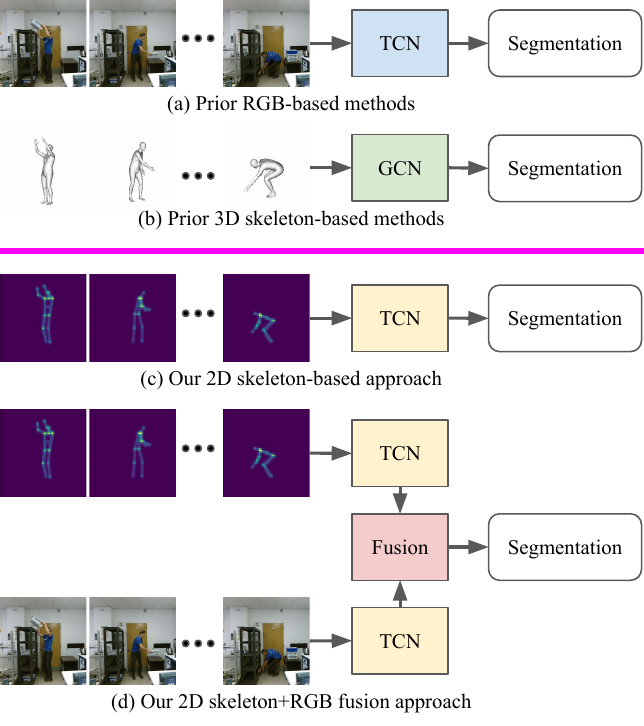}
    \caption{Prior methods either take sequences of RGB frames (a) or sequences of 3D skeletons (b) as inputs. We propose a new approach which relies on sequences of 2D skeleton heatmaps (c). We further explore 2D skeleton+RGB fusion (d) for action segmentation, leading to performance gains.}
    \label{fig:teaser}
\end{figure}

A majority of action segmentation methods, e.g., \cite{lea2017temporal,ding2017tricornet,lei2018temporal,farha2019ms,li2020ms}, take RGB videos as inputs, which are then passed through TCNs to capture long-term action dependencies and predict segmentation results (see Fig.~\ref{fig:teaser}(a)). Since the introduction of TCNs to action segmentation in the pioneering work of Lea et al.~\cite{lea2017temporal}, several improvements have been proposed. For example, multi-stage TCNs~\cite{farha2019ms,li2020ms} operate on the full temporal resolution of the videos, as compared to downsampling the temporal resolution of the videos as in Lea et al.~\cite{lea2017temporal}.

In contrast with the above RGB-based methods, skeleton-based alternatives~\cite{parsa2020spatio,parsa2021multi,filtjens2022skeleton} have attracted research interests only recently because of their action focus and compact representation. These skeleton-based methods~\cite{parsa2020spatio,parsa2021multi,filtjens2022skeleton} take 3D skeleton sequences as inputs and employ GCNs to process 3D skeleton sequences directly due to their irregular graph structures (see Fig.~\ref{fig:teaser}(b)). As a result, it is difficult to incorporate other modalities that have regular grid structures (e.g., RGB, depth, flow) into existing skeleton-based methods~\cite{parsa2020spatio,parsa2021multi,filtjens2022skeleton}.

In this work, we present a novel skeleton-based action segmentation approach, which relies only on 2D skeleton sequences. Inspired by the success of Duan et al.~\cite{duan2022revisiting} in skeleton-based action recognition, we transform sequences of 2D skeletons into sequences of heatmaps. Since our heatmaps have image-like structures, i.e., $W \times H \times C$ with $W = H = 56$, $C = 3$, we can extract features from the heatmaps by using pre-trained ResNet~\cite{he2016deep}/VGG~\cite{simonyan2014very} before feeding the features to TCNs for action segmentation (see Fig.~\ref{fig:teaser}(c)), in a similar manner as employing multi-stage TCNs~\cite{farha2019ms,li2020ms} on RGB videos. Experiments on action segmentation datasets demonstrate that our 2D skeleton-based approach achieves similar/better performances and higher robustness against missing keypoints than previous 3D skeleton-based methods. In addition, we boost the performances further via multi-modality fusion, i.e., by combining RGB inputs with 2D skeleton inputs (see Fig.~\ref{fig:teaser}(d)). More specifically, we introduce fusion modules at multiple stages of~\cite{farha2019ms,li2020ms} to facilitate deep supervision~\cite{lee2015deeply,li2017deep,li2018deep}.

In summary, our contributions include:
\begin{itemize}
    \item We propose a 2D skeleton-based action segmentation method, which takes 2D skeleton heatmaps as inputs and utilizes TCNs to capture spatiotemporal features. Our approach achieves comparable/better results and higher robustness against missing keypoints than 3D skeleton-based methods which operate directly on 3D skeleton coordinates and employ GCNs for spatiotemporal feature extraction.
    \item We further fuse 2D skeleton heatmaps with RGB videos, yielding improved performances. To the best of our knowledge, our work is the first to utilize 2D skeleton heatmap inputs and the first to explore 2D skeleton+RGB fusion for action segmentation.
    \item We annotate framewise action labels and obtain estimated 2D/3D skeletons for TUM-Kitchen, which are available at \url{https://bitly.ws/3eWhV}.
\end{itemize}
\section{Related Work}
\label{sec:relatedwork}

%In the following, we discuss recent literature on action segmentation, including both RGB-based and skeleton-based approaches, and skeleton-based action recognition.

\noindent \textbf{RGB-Based Action Segmentation.}
RGB-based methods, e.g., \cite{lea2017temporal,ding2017tricornet,lei2018temporal,farha2019ms,li2020ms}, typically employ TCNs to learn long-term action dependencies. TCNs are first applied in Lea at al.~\cite{lea2017temporal}, which perform temporal convolutions and deconvolutions. TricorNet~\cite{ding2017tricornet} further utilizes bi-directional LSTMs, while TDRN~\cite{lei2018temporal} uses deformable temporal convolutions instead. One drawback of the above methods is that they temporally downsample videos. To address that, multi-stage TCNs~\cite{farha2019ms,li2020ms} are designed to preserve temporal resolutions. Recently, refinement techniques, e.g., \cite{ishikawa2021alleviating,huang2020improving}, are developed to reduce over-segmentations. Instead of RGB inputs, in this paper we use 2D skeleton inputs and introduce 2D skeleton+RGB fusion for action segmentation.

\noindent \textbf{Skeleton-Based Action Segmentation.}
Skeleton-based approaches~\cite{parsa2020spatio,parsa2021multi,filtjens2022skeleton} have emerged only recently. In particular, Parsa et al.~\cite{parsa2020spatio} propose a GCN architecture for skeleton-based action segmentation, while a GCN backbone is employed in their succeeding work~\cite{parsa2021multi} for joint skeleton-based action segmentation and ergonomics analysis. Recently, Filtjens et al.~\cite{filtjens2022skeleton} extend multi-stage TCNs~\cite{farha2019ms,li2020ms} for RGB inputs to multi-stage GCNs for skeleton inputs. All of the above methods use 3D skeleton inputs and GCNs, which makes it hard to combine with other modalities (e.g., RGB, depth, flow). In contrast, our approach employs 2D skeleton inputs and TCNs, yielding comparable/superior results and better robustness against missing keypoints. Further, our 2D skeleton+RGB fused version leads to performance gains. Note that 2D skeletons have been used in previous methods~\cite{kobayashi2019fine,ma2021fine}. However, Kobayashi et al.~\cite{kobayashi2019fine} compute hand heatmaps for reweighting image features only, while Ma et al.~\cite{ma2021fine} operate directly on 2D skeleton coordinates. 

\noindent \textbf{Skeleton-Based Action Recognition.}
Action recognition with skeleton inputs can roughly be grouped into GCN-based methods, e.g., \cite{yan2018spatial,song2020stronger,cai2021jolo,chen2021channel}, and CNN-based ones, e.g., \cite{liu2017two,caetano2019skelemotion,lin2020image,asghari2020dynamic}. The former process skeleton sequences directly with GCNs, whereas the latter aggregate them into regular grid inputs that are suitable for CNNs. While GCN-based methods are difficult to fuse with other modalities  (e.g., RGB, depth, flow), CNN-based ones suffer from information loss during aggregation. To address the above, Duan et al.~\cite{duan2022revisiting} model skeleton sequences as heatmap sequences and pass the heatmap sequences to CNNs, yielding superior results. Motivated by Duan et al.~\cite{duan2022revisiting}, we exploit heatmap representations and heatmap+RGB fusion for the fine-grained action segmentation task.
\section{Our Approach}
\label{sec:method}

Below we present our main contributions, including a 2D skeleton-based approach and a multi-modality approach with both RGB and 2D skeleton inputs for action segmentation.

\subsection{2D Skeleton Heatmap}
\label{sec:heatmap}

\begin{figure}[t]
    \centering
	\includegraphics[width=0.9\linewidth, trim = 0mm 70mm 40mm 0mm, clip]{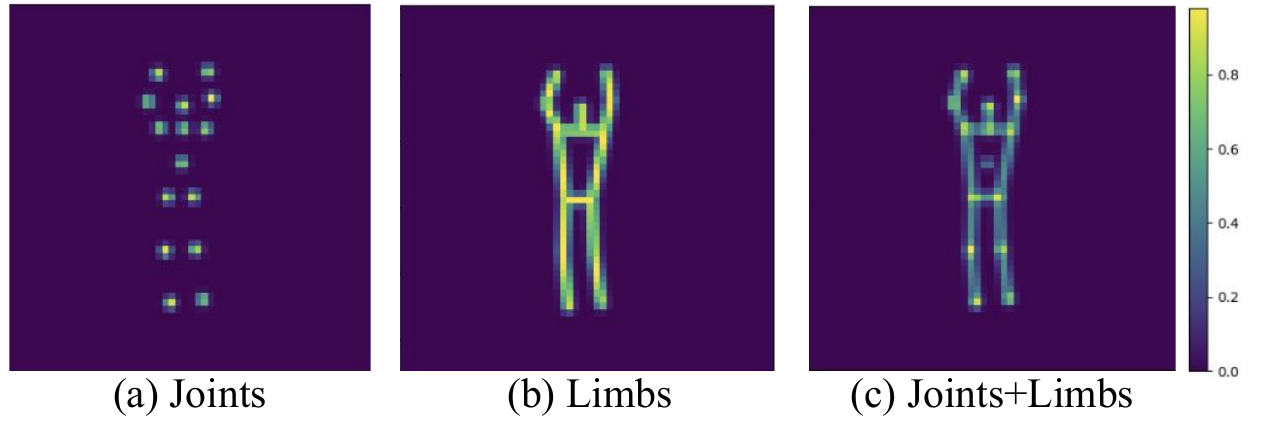}
    \caption{Examples of 2D skeleton heatmaps.}
    \label{fig:heatmap}
\end{figure}

\begin{figure*}[t]
    \centering
	\includegraphics[width=0.9\linewidth, trim = 0mm 30mm 10mm 0mm, clip]{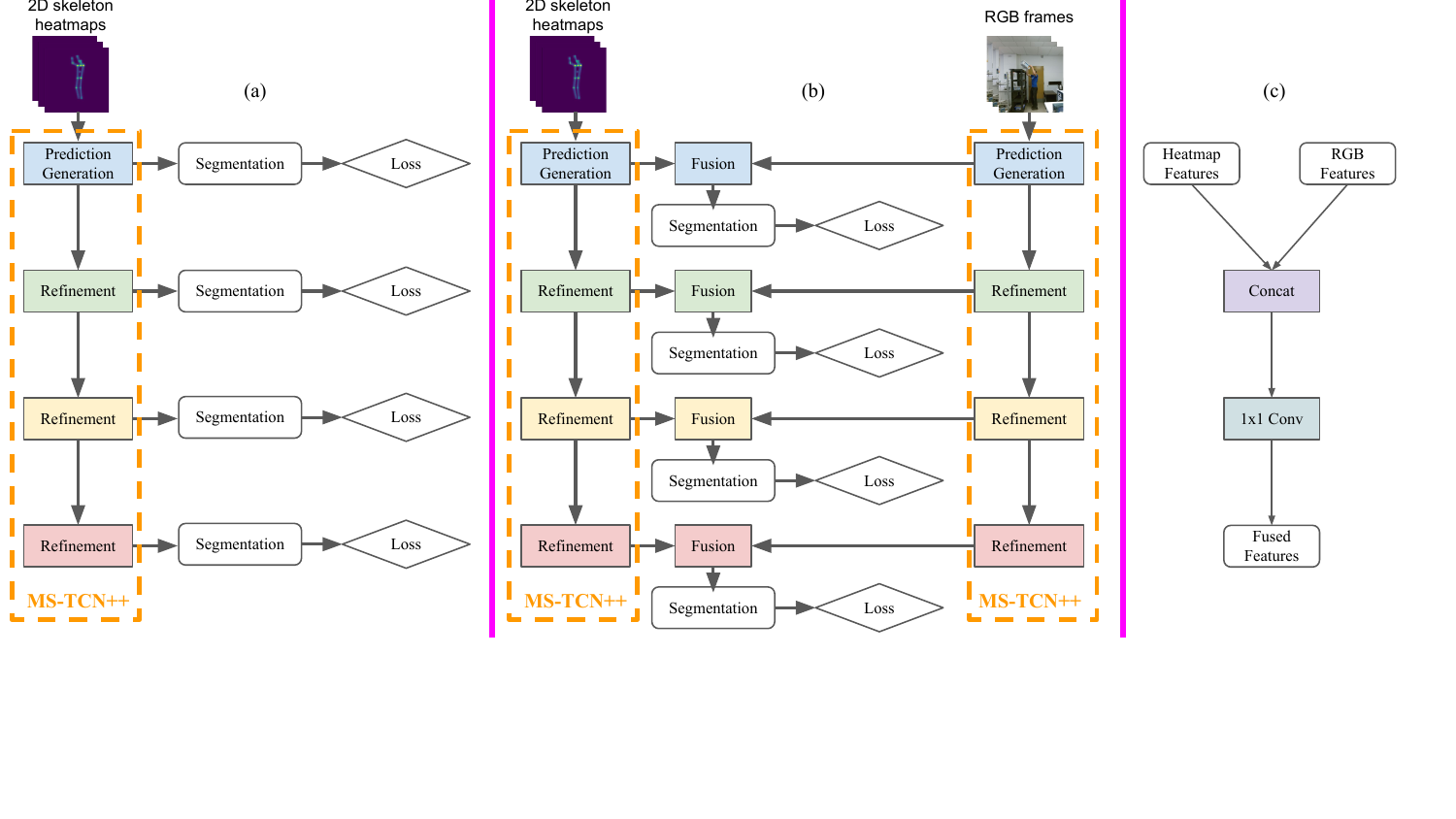}
    \caption{(a) 2D skeleton-based action segmentation. We convert 2D skeletons into image-like heatmaps, which are passed to an RGB-based network for action segmentation, i.e., MS-TCNN++~\cite{li2020ms}. (b) 2D skeleton+RGB-based action segmentation. During training, we propose fusion modules at various stages of MS-TCN++~\cite{li2020ms} for deep supervision~\cite{lee2015deeply,li2018deep}. At testing, the segmentation predicted by the last refinement stage is considered as our output. (c) 2D skeleton+RGB fusion module.}
    \label{fig:method}
\end{figure*}

Given video frames $\boldsymbol{F}$, a well-known human detector, e.g., Faster-RCNN~\cite{girshick2015fast}, and a recent top-down 2D human pose estimator, e.g., HRNet~\cite{sun2019deep}, can first be applied for extracting 2D skeletons $\boldsymbol{S}$ with high quality. Here, we model a 2D skeleton $\boldsymbol{s}$ by a set of 2D joint triplets $\{(x_k,y_k,c_k)\}$, where the $k$-th joint has 2D coordinates $(x_k,y_k)$ and (maximum) confidence score $c_k$. In the following, 2D skeletons $\boldsymbol{S}$ are transformed into heatmaps $\boldsymbol{H}$. For a 2D skeleton $\boldsymbol{s}$, we derive a heatmap $\boldsymbol{h}$ of size $W \times H \times K$ (with the width $W$ and height $H$ of the video frame and the number of 2D joints $K$). Specifically, given a set of 2D joint triplets $\{(x_k,y_k,c_k)\}$, we derive a \emph{joint} heatmap $\boldsymbol{h}^J$, which includes $K$ Gaussian distributions centered at every 2D joint, as:
\begin{align}
    \boldsymbol{h}^J_{ijk} = e^{-\frac{(i-x_k)^2+(j-y_k)^2}{2*\sigma^2}} * c_k,
\end{align}
with standard deviation $\sigma = 0.6$. Alternatively, we derive a \emph{limb} heatmap $\boldsymbol{h}^L$ of size $W \times H \times L$ (with the number of limbs $L$) as:
\begin{align}
    \boldsymbol{h}^{L}_{ijl} = e^{-\frac{dist((i,j), seg(a_l,b_l))}{2*\sigma^2}} * min(c_{a_l},c_{b_l}).
\end{align}
Here, the $l$-th limb denotes the segment $seg(a_l,b_l)$ between the joints $a_l$ and $b_l$, and the $dist$ function denotes the distance from the location $(i,j)$ to the segment $seg(a_l,b_l)$. As we will show in Sec.~\ref{sec:exp_heatmap}, combining joint and limb heatmaps yields the best results. Thus, for each 2D skeleton $\boldsymbol{s}_i$, we derive the \emph{combined} heatmap $\boldsymbol{h}_i = \boldsymbol{h}^{J+L}_i$. In contrast with all previous skeleton-based methods~\cite{parsa2020spatio,parsa2021multi,filtjens2022skeleton}, which use 3D skeletons, we rely only on 2D skeletons. Also, accurate 2D skeletons are easier to obtain than accurate 3D skeletons, as accurate depths are not required. Lastly, as we will show in Sec.~\ref{sec:exp_robustness}, 2D skeleton heatmaps are more robust than 3D skeleton coordinates as joints/limbs are represented by Gaussian distributions.

In addition, we crop $\boldsymbol{h}_i$ along the spatial dimensions by using the smallest bounding box containing all 2D skeletons in the entire video sequence. Next, we sum all component (joint/limb) heatmaps within $\boldsymbol{h}_i$ into a single heatmap and resize it to $56 \times 56$. Finally, we replicate the heatmap and stack the copies, yielding an image-like heatmap $\boldsymbol{h}_i$ of size $56 \times 56 \times 3$, which can be used for extracting pre-trained features. Fig.~\ref{fig:heatmap} illustrates examples of 2D skeleton heatmaps. Unlike Duan et al.~\cite{duan2022revisiting}, which are mainly interested in video-level cues for coarse-grained action recognition (predicting a \emph{single} label for a video) and perform temporal sampling to reduce computation costs, our work addresses fine-grained action segmentation (predicting \emph{framewise} labels) where frame-level cues are important and hence temporal sampling is not employed in our method.

\subsection{2D Skeleton-Based Action Segmentation}
\label{sec:segmentation}

As described above, our 2D skeleton heatmaps resemble RGB images. Thus, we can extract features from the 2D skeleton heatmaps by using ResNet~\cite{he2016deep}/VGG~\cite{simonyan2014very} pre-trained on ImageNet~\cite{deng2009imagenet}. Next, an RGB-based action segmentation network can be employed to perform action segmentation on the extracted features. Here, we choose MS-TCN++~\cite{li2020ms} due to its accuracy and efficiency. Fig.~\ref{fig:method}(a) shows an overview of our 2D skeleton-based action segmentation method. The network includes one prediction generation stage, which computes the initial segmentation, and three refinement stages, which improve the initial segmentation. The prediction generation stage consists of eleven dilated temporal convolutional layers, while the refinement stages share the same ten dilated temporal convolutional layers.

To train MS-TCN++~\cite{li2020ms}, deep supervision~\cite{lee2015deeply,li2018deep} is employed. Particularly, all four stages are trained with the same combination of classification loss and smoothness loss. The classification loss is computed as the cross-entropy loss:
\begin{align}
    \mathcal{L}_{class} = \frac{1}{M} \sum_{i} -\log y_{i,c},
\end{align}
with the number of video frames $M$ and the probability $y_{i,c}$ of assigning frame $\boldsymbol{f}_i$ to ground truth action class $c$. To reduce over-segmentation, the smoothness loss is added:
\begin{align}
    \mathcal{L}_{smooth} = \frac{1}{MC} \sum_{i,c} \Tilde{\Delta}^2_{i,c},\\
    \Tilde{\Delta}_{i,c} = \begin{cases}
        \Delta_{i,c}, & \Delta_{i,c} \leq \tau\\
        \tau, & \Delta_{i,c} > \tau\\
    \end{cases},\\
    \Delta_{i,c} = \left| \log y_{i,c} - \log y_{i-1,c} \right|,
\end{align}
with the number of action classes $C$ and thresholding parameter $\tau = 16$. The final loss is written as:
\begin{align}
    \mathcal{L} = \mathcal{L}_{class} + \alpha \mathcal{L}_{smooth},  
    \label{eq:final_loss}
\end{align}
with balancing parameter $\alpha = 0.15$. At testing, the prediction by the last refinement stage is used as our output.

\subsection{2D Skeleton+RGB-Based Action Segmentation}
\label{sec:fusion}

In the following, we explore multi-modality fusion to improve the performance. Fig.~\ref{fig:method}(b) shows an overview of our 2D skeleton+RGB-based action segmentation method. The network consists of two branches for processing 2D skeleton heatmaps and RGB frames respectively. Since our 2D skeleton heatmaps have the same structures as the RGB frames, we use the same MS-TCN++~\cite{li2020ms} architecture for both branches. To perform 2D skeleton+RGB fusion, we introduce fusion modules at all stages from prediction generation stage to refinement stages. In particular, our fusion module first concatenates the heatmap feature vector with the RGB feature vector before passing the concatenated feature vector through a $1 \times 1$ convolutional layer to reduce the fused feature vector to the original size of 64. Fig.~\ref{fig:method}(c) illustrates the above steps. We follow MS-TCN~\cite{li2020ms} to use the same losses as described in the previous section. At testing, our output is the prediction by the last refinement stage.

In addition, we have experimented with adding augmentation to 2D skeleton heatmaps (i.e., temporal augmentation, position augmentation, orientation augmentation, and horizontal flipping) and RGB frames (i.e., temporal augmentation, brightness augmentation, contrast augmentation, and horizontal flipping). However, the performance gain is marginal while the computational cost is increased significantly. Therefore, we do not apply data augmentation in this work. In contrast with Parsa et al.~\cite{parsa2020spatio}, which operate directly on 3D skeleton coordinates, our 2D skeleton heatmap representation makes it easier to fuse with RGB frames. Furthermore, while Parsa et al.~\cite{parsa2020spatio} conduct fusion at the final level only, we perform fusion across multiple levels.
\section{Experiments}
\label{sec:experiments}

%In this section, we benchmark our 2D skeleton-based approach and our 2D skeleton+RGB-based approach against state-of-the-art methods, including those with 3D skeleton inputs and RGB inputs, on action segmentation datasets.

\noindent \textbf{Datasets, Annotations, and Poses.} We use three datasets, which capture a diverse set of human activities:
\begin{itemize}
    \item \emph{UW-IOM}~\cite{parsa2019toward} includes 20 videos of a warehouse activity, which consists of 17 actions (e.g., \emph{``box\_bend\_pick\_up\_low''}). Each video is $\sim$3 minutes long. We use both framewise action labels and 2D/3D human poses (estimated with~\cite{rogez2017lcr}) released by~\cite{parsa2021multi}.
    \item \emph{TUM-Kitchen}~\cite{tenorth2009tum} consists of 19 videos of a kitchen activity, which comprises of 21 actions (e.g., \emph{``walk,hold-both-hand''}). The duration of each video is $\sim$2 minutes. We manually annotate framewise action labels and obtain 2D/3D human poses with~\cite{rogez2017lcr}, as they are not provided by~\cite{parsa2021multi}.
    \item \emph{Desktop Assembly}~\cite{kumar2022unsupervised,tran2024permutation} includes 76 videos of an assembly activity. The activity consists of 23 actions (e.g., \emph{``tighten\_screw\_4''}) and the video length is $\sim$1.5 minutes. We use framewise action labels of~\cite{kumar2022unsupervised}, while estimating 2D human poses with~\cite{sun2019deep} and 3D human poses with~\cite{li2022mhformer}.
\end{itemize}

\noindent \textbf{Implementation Details.} Our heatmap-based model is implemented in pyTorch. We randomly initialize our model and use ADAM optimization with a learning rate of $0.001$. We train our model for 100 epochs. Furthermore, our heatmap+RGB-based model is trained in two stages: i) we first train the heatmap and RGB branches separately with the same losses in Eq.~\ref{eq:final_loss}, ii) we then train our entire fusion model (with a learning rate of $0.0005$) by using the weights from the first stage as initialization.

\noindent \textbf{Competing Methods.} We test against prior skeleton-based methods, which rely on 3D skeletons and GCNs. They include ST-PGN~\cite{parsa2020spatio}, MTL/STL~\cite{parsa2021multi}, and MS-GCN~\cite{filtjens2022skeleton}. Moreover, we evaluate the classical MS-TCN++~\cite{li2020ms} with RGB inputs. Lastly, we compare with the 3D skeleton+RGB fusion version of ST-PGN~\cite{parsa2020spatio}. If the results are reported in the original papers, we copy them (except for TUM-Kitchen, since we use different sets of labels/poses). Otherwise, we run the official implementations~\footnote{MTL/STL: \url{https://github.com/BehnooshParsa/MTL-ERA}\\ MS-GCN: \url{https://github.com/BenjaminFiltjens/MS-GCN}\\ MS-TCN++: \url{https://github.com/sj-li/MS-TCN2}} to obtain the results.

\noindent \textbf{Metrics.} Following~\cite{parsa2020spatio,parsa2021multi}, we use \emph{F1} score (10\% overlap), \emph{Edit} distance, and mean Average Precision (\emph{mAP}). In addition, we follow~\cite{filtjens2022skeleton,li2020ms} to add framewise Accuracy (\emph{Acc}). Please refer to~\cite{parsa2020spatio,parsa2021multi,filtjens2022skeleton,li2020ms} for detailed definitions.

\subsection{Impacts of Different Heatmaps}
\label{sec:exp_heatmap}

\begin{table*}[t]
\begin{minipage}[b]{1.0\linewidth}
\centering

{%
\setlength{\tabcolsep}{2pt}
\begin{tabular}{c|c|c|c|c}

\specialrule{1pt}{1pt}{1pt}

\textbf{\small{Heatmap}} & \textbf{\small{F1}} & \textbf{\small{Edit}}  & \textbf{\small{mAP}} & \textbf{\small{Acc}}  \\
\midrule
Joint&89.87 $\pm$ 05.95&89.77 $\pm$ 04.86&85.07 $\pm$ 06.71 &79.62 $\pm$ 06.93  \\
Limb&\underline{\textit{93.41 $\pm$ 02.85}}&\underline{\textit{92.20 $\pm$ 03.92}}&\underline{\textit{88.63 $\pm$ 03.66}}  &\underline{\textit{83.15 $\pm$ 03.37}}  \\
\cellcolor{beaublue}Joint+Limb&\cellcolor{beaublue}\textbf{93.82 $\pm$ 02.90}&\cellcolor{beaublue}\textbf{92.88 $\pm$ 04.13}&\cellcolor{beaublue}\textbf{89.82 $\pm$ 03.83}  &\cellcolor{beaublue}\textbf{85.04 $\pm$ 03.08}  \\

\specialrule{1pt}{1pt}{1pt}
\end{tabular}

}

\caption{Impacts of different heatmaps on UW-IOM. Best results are in \textbf{bold}. Second best results are \underline{\textit{underlined}}.}

\label{tab:ablation-heatmap}
\end{minipage}

\end{table*}

We first study the performance of our 2D skeleton-based approach by using various inputs: joint heatmaps, limb heatmaps and integrated joint+limb heatmaps. We use the UW-IOM dataset and ResNet-50 features in this experiment. The ablation results are shown in Tab.~\ref{tab:ablation-heatmap}. It is evident that the best performance is achieved by combined joint+limb heatmaps, followed by limb heatmaps alone, whereas the worst performance is obtained by joint heatmaps alone.

\subsection{Impacts of Different Features}
\label{sec:exp_feature}

\begin{table*}[t]
\begin{minipage}[b]{1.0\linewidth}
\centering

{%
\setlength{\tabcolsep}{2pt}
\begin{tabular}{c|c|c|c|c}

\specialrule{1pt}{1pt}{1pt}

\textbf{\small{Feature}} & \textbf{\small{F1}} & \textbf{\small{Edit}}  & \textbf{\small{mAP}} & \textbf{\small{Acc}}  \\
\midrule
VGG-16&91.58 $\pm$ 04.38&92.26 $\pm$ 04.91&87.20 $\pm$ 02.98  &79.42 $\pm$ 04.22  \\
ResNet-18&\underline{\textit{93.76 $\pm$ 03.23}}&\underline{\textit{92.79 $\pm$ 04.24}}&\underline{\textit{88.01 $\pm$ 04.19}}&\underline{\textit{84.53 $\pm$ 04.07}}\\
\cellcolor{beaublue}ResNet-50&\cellcolor{beaublue}\textbf{93.82 $\pm$ 02.90}&\cellcolor{beaublue}\textbf{92.88 $\pm$ 04.13}&\cellcolor{beaublue}\textbf{89.82 $\pm$ 03.83}  &\cellcolor{beaublue}\textbf{85.04 $\pm$ 03.08}\\

\specialrule{1pt}{1pt}{1pt}
\end{tabular}

}

\caption{Impacts of different features on UW-IOM. Best results are in \textbf{bold}. Second best results are \underline{\textit{underlined}}.}

\label{tab:ablation-feature}
\end{minipage}

\end{table*}

We now investigate the performance of our 2D skeleton-based method by using various networks pre-trained on ImageNet for feature extraction: VGG-16, ResNet-18, and ResNet-50. The UW-IOM dataset and joint+limb heatmaps are used in this experiment. Tab.~\ref{tab:ablation-feature} presents the ablation results. It is clear that ResNet-50 features yield the best results, outperforming ResNet-18 features and VGG-16 features.

\subsection{Robustness against Missing Keypoints}
\label{sec:exp_robustness}

\begin{table*}[t]
\begin{minipage}[b]{1.0\linewidth}
\centering

{%
\setlength{\tabcolsep}{2pt}
\begin{tabular}{c|c|c|c|c|c}

\specialrule{1pt}{1pt}{1pt}

\textbf{\small{Method}} & \textbf{\small{Input}}& \multicolumn{4}{c}{\textbf{\small{Keypoint Missing Probability $p$}}} \\ 
 \cline{3-6}
&&\textbf{\small{0\%}}&\textbf{\small{25\%}}&\textbf{\small{50\%}}&\textbf{\small{100\%}}\\
\midrule
STL~\cite{parsa2021multi} &$\triangle$& 87.24 $\pm$ 01.50 & 86.19 $\pm$ 02.27 & 84.29 $\pm$ 03.71 & 80.25 $\pm$ 05.68 \\
MS-GCN~\cite{filtjens2022skeleton} &$\triangle$& \underline{\textit{91.93 $\pm$ 04.60}} &\underline{\textit{90.14 $\pm$ 04.63}}&\underline{\textit{89.33 $\pm$ 04.83}}  &\underline{\textit{88.05 $\pm$ 04.64}}  \\
\cellcolor{beaublue} Ours &\cellcolor{beaublue}$\star$ &\cellcolor{beaublue}\textbf{93.82 $\pm$ 02.90} &\cellcolor{beaublue}\textbf{93.48 $\pm$ 02.95} &\cellcolor{beaublue}\textbf{93.43 $\pm$ 02.75} &\cellcolor{beaublue}\textbf{91.28 $\pm$ 02.65} \\

\specialrule{1pt}{1pt}{1pt}
\end{tabular}

}

\caption{Robustness against missing keypoints on UW-IOM. Note that $\triangle$ denotes 3D pose coordinate inputs and $\star$ denotes 2D pose heatmap inputs. Best results are in \textbf{bold}. Second best results are \underline{\textit{underlined}}.}

\label{tab:ablation-missing}
\end{minipage}

\end{table*}

\begin{figure}[t]
    \centering
	\includegraphics[width=0.9\linewidth, trim = 0mm 70mm 40mm 0mm, clip]{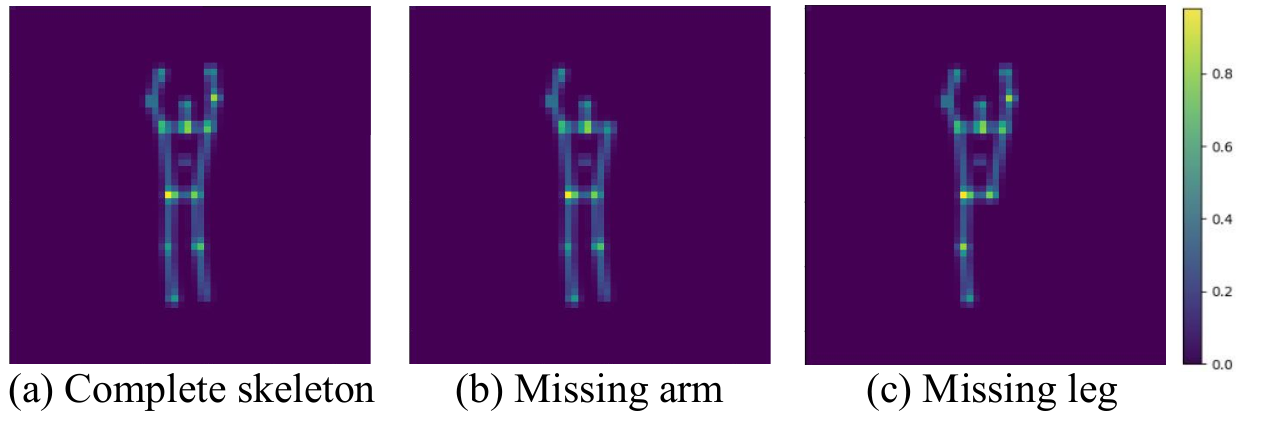}
    \caption{Examples of missing keypoints.}
    \label{fig:missing}
\end{figure}

We conduct an experiment in which a certain component of the skeleton is removed to examine the robustness of our 2D skeleton-based approach. In particular, at testing (without any finetuning), we randomly drop a limb (i.e., arm or leg) with a probability $p$ for each frame in the UW-IOM dataset. Fig.~\ref{fig:missing} shows examples of missing keypoints. The F1 scores of our method, MS-GCN~\cite{filtjens2022skeleton}, and STL~\cite{parsa2021multi} are illustrated in Tab.~\ref{tab:ablation-missing}. It can be seen that our approach demonstrates a high level of robustness against missing keypoints. For example, when one limb is dropped per frame (i.e., $p = 100\%$), our F1 score drop is $2.54\%$, as compared to $3.88\%$ and $6.99\%$ of MS-GCN~\cite{filtjens2022skeleton} and STL~\cite{parsa2021multi} respectively. Moreover, our F1 score remains stable for $p = 25\%$ and $p = 50\%$, whereas the F1 scores of MS-GCN~\cite{filtjens2022skeleton} and STL~\cite{parsa2021multi} decrease up to $2.60\%$ and $2.95\%$ respectively. This is likely because our method models each joint as a Gaussian and uses TCNs for learning spatiotemporal features, whereas MS-GCN~\cite{filtjens2022skeleton} and STL~\cite{parsa2021multi} represent each joint by its coordinates and use GCNs for spatiotemporal feature extraction.

\subsection{Comparisons on UW-IOM}
\label{sec:exp_uw}

\begin{table*}[t]
\begin{minipage}[t]{1.0\linewidth}
\centering

{%
\setlength{\tabcolsep}{2pt}
\begin{tabular}{c|c|c|c|c|c}

\specialrule{1pt}{1pt}{1pt}

 \textbf{\small{Method}} & \textbf{\small{Input}} & \textbf{\small{F1}} & \textbf{\small{Edit}} & \textbf{\small{mAP}} & \textbf{\small{Acc}}  \\
\midrule
MS-TCN++~\cite{li2020ms}&$\square$&93.36 $\pm$ 03.35&92.17 $\pm$ 04.08&87.99 $\pm$ 05.10  &82.29 $\pm$ 03.89  \\
$^\dagger$ST-PGN~\cite{parsa2020spatio}&$\triangle$&87.95 $\pm$ 01.54&\textbf{97.86 $\pm$ 02.15}& 87.03 $\pm$ 02.85 & - \\
$^\dagger$ST-PGN (Fusion)~\cite{parsa2020spatio}&$\triangle,\square$&88.08 $\pm$ 01.89&80.90 $\pm$ 02.06&87.05 $\pm$ 03.47  & - \\
$^\dagger$MTL~\cite{parsa2021multi}&$\triangle$&92.03 $\pm$ 02.54&91.59 $\pm$ 01.23&74.45 $\pm$ 10.36  & - \\
$^\dagger$STL~\cite{parsa2021multi}&$\triangle$&92.33 $\pm$ 00.78&92.08 $\pm$ 01.18&49.61 $\pm$ 00.17  & - \\
MS-GCN~\cite{filtjens2022skeleton}&$\triangle$&91.93 $\pm$ 04.60&87.61 $\pm$ 05.87&87.52 $\pm$ 05.29  &82.75 $\pm$ 05.04  \\
\cellcolor{beaublue} Ours &\cellcolor{beaublue}$\star$ &\cellcolor{beaublue}\underline{\textit{93.82 $\pm$ 02.90}}&\cellcolor{beaublue}92.88 $\pm$ 04.13&\cellcolor{beaublue}\underline{\textit{89.82 $\pm$ 03.83}}&\cellcolor{beaublue}\underline{\textit{85.04 $\pm$ 03.08}}   \\
\cellcolor{beaublue} Ours (Fusion)&\cellcolor{beaublue}$\star,\square$&\cellcolor{beaublue}\textbf{94.54 $\pm$ 02.75}&\cellcolor{beaublue}\underline{\textit{93.25 $\pm$ 03.81}}&\cellcolor{beaublue}\textbf{90.12 $\pm$ 03.65}  &\cellcolor{beaublue}\textbf{85.84 $\pm$ 03.27}  \\

\specialrule{1pt}{1pt}{1pt}
\end{tabular}

}

\caption{Quantitative comparisons on UW-IOM. Note that $\square$ denotes direct RGB inputs, $\triangle$ denotes 3D pose coordinate inputs, and $\star$ denotes 2D pose heatmap inputs. Also, $^\dagger$ indicates that results are copied from the original papers. Best results are in \textbf{bold}. Second best results are \underline{\textit{underlined}}.}
\label{tab:comparison-uw}
\end{minipage}

\end{table*}

%\begin{figure*}[t]
%    \centering
%	\includegraphics[width=0.95\linewidth, trim = 0mm 100mm 0mm 0mm, clip]{Figures/uw.pdf}
%    \caption{Qualitative comparisons on UW-IOM (sequence \emph{10}).}
%    \label{fig:uw}
%\end{figure*}

In this section, we compare our 2D skeleton-based and 2D skeleton+RGB-based approaches with state-of-the-art action segmentation methods, including ST-PGN~\cite{parsa2020spatio}, MTL/STL~\cite{parsa2021multi}, MS-GCN~\cite{filtjens2022skeleton}, and MS-TCN++~\cite{li2020ms}, on UW-IOM. The quantitative results are shown in Tab.~\ref{tab:comparison-uw}. It is evident that our 2D skeleton+RGB fusion approach achieves the best overall results, outperforming the 3D skeleton+RGB fusion version of ST-PGN~\cite{parsa2020spatio} by large margins on F1 score, Edit distance, and mAP, e.g., $92.88\%$ Edit distance for Ours (Fusion) vs. $80.90\%$ for ST-PGN (Fusion). Next, our 2D skeleton-based approach obtains the second best overall results, outperforming previous 3D skeleton-based and RGB-based methods on F1 score, mAP, and Acc. The results in Tab.~\ref{tab:comparison-uw} confirm the effectiveness of using 2D skeleton heatmaps and TCNs for capturing spatiotemporal features. %Lastly, we include some qualitative results in Fig.~\ref{fig:uw}, where our results closely match the ground truth. 

\subsection{Comparisons on TUM-Kitchen}
\label{sec:exp_tum}

\begin{table*}[t]
\begin{minipage}[t]{1.0\linewidth}
\centering

{%
\setlength{\tabcolsep}{2pt}
\begin{tabular}{c|c|c|c|c|c}

\specialrule{1pt}{1pt}{1pt}

\textbf{\small{Method}} & \textbf{\small{Input}} & \textbf{\small{F1}} & \textbf{\small{Edit}} & \textbf{\small{mAP}} & \textbf{\small{Acc}}  \\
\midrule
MS-TCN++~\cite{li2020ms}&$\square$& 81.75 $\pm$ 04.04 & 84.76 $\pm$ 02.90 & 56.44 $\pm$ 02.63 & 69.28 $\pm$ 03.98 \\
STL~\cite{parsa2021multi}&$\square$& 78.81 $\pm$ 08.42 & 81.50 $\pm$ 07.57 & 46.24 $\pm$ 17.42 & 59.77 $\pm$ 15.76 \\
MS-GCN~\cite{filtjens2022skeleton}&$\triangle$& 76.30 $\pm$ 04.23 & 80.14 $\pm$ 03.36 & 57.20 $\pm$ 02.59 & 69.38 $\pm$ 03.70 \\
\cellcolor{beaublue} Ours &\cellcolor{beaublue}$\star$ &\cellcolor{beaublue}\textbf{81.96 $\pm$ 03.72}
 &\cellcolor{beaublue}\underline{\textit{85.14 $\pm$ 03.01}} &\cellcolor{beaublue}\underline{\textit{58.81 $\pm$ 03.95}} &\cellcolor{beaublue}\underline{\textit{71.55 $\pm$ 04.75}} \\
\cellcolor{beaublue} Ours (Fusion) &\cellcolor{beaublue}$\star,\square$&\cellcolor{beaublue}\underline{\textit{81.38 $\pm$ 02.86}}  &\cellcolor{beaublue}\textbf{85.33 $\pm$ 02.16}
 &\cellcolor{beaublue}\textbf{61.40 $\pm$ 01.93} &\cellcolor{beaublue}\textbf{72.89 $\pm$ 03.77} \\

\specialrule{1pt}{1pt}{1pt}
\end{tabular}

}

\caption{Quantitative comparisons on TUM-Kitchen. Note that $\square$ denotes direct RGB inputs, $\triangle$ denotes 3D pose coordinate inputs, and $\star$ denotes 2D pose heatmap inputs. Best results are in \textbf{bold}. Second best results are \underline{\textit{underlined}}.}
\label{tab:comparison-tum}
\end{minipage}

\end{table*}

%\begin{figure*}[t]
%    \centering
%	\includegraphics[width=0.95\linewidth, trim = 0mm 100mm 0mm 0mm, clip]{Figures/tum.pdf}
%    \caption{Qualitative comparisons on TUM-Kitchen (sequence \emph{10}).}
%    \label{fig:tum}
%\end{figure*}

Tab.~\ref{tab:comparison-tum} presents the quantitative results on TUM-Kitchen. It is clear that our 2D skeleton-based approach outperforms previous 3D skeleton-based and RGB-based methods on all metrics, e.g., on Acc, $71.55\%$ for Ours, as compared to $69.38\%$, $59.77\%$, and $69.28\%$ for MS-GCN~\cite{filtjens2022skeleton}, STL~\cite{parsa2021multi}, and MS-TCN++~\cite{li2020ms} respectively. Further, fusing 2D skeleton inputs and RGB inputs leads to the best overall performance with the best numbers on Edit distance, mAP, and Acc. The above observations validate the use of 2D skeleton heatmaps for action segmentation. %Finally, some qualitative results are illustrated in Fig.~\ref{fig:tum}, where our results are more closely aligned with the ground truth than MS-GCN~\cite{filtjens2022skeleton} and MS-TCN++~\cite{li2020ms}.

\subsection{Comparisons on Desktop Assembly}
\label{sec:exp_da}

\begin{table*}[t]
\begin{minipage}[t]{1.0\linewidth}
\centering

{%
\setlength{\tabcolsep}{2pt}
\begin{tabular}{c|c|c|c|c|c}

\specialrule{1pt}{1pt}{1pt}

\textbf{\small{Method}} & \textbf{\small{Input}} & \textbf{\small{F1}} & \textbf{\small{Edit}} & \textbf{\small{mAP}} & \textbf{\small{Acc}}  \\
\midrule
MS-TCN++~\cite{li2020ms}&$\square$&97.24 $\pm$ 02.08&\textbf{98.05 $\pm$ 02.11}&91.00 $\pm$ 04.62 &87.42 $\pm$ 03.35  \\
STL~\cite{parsa2021multi}&$\triangle$& 87.16 $\pm$ 07.23& 85.71 $\pm$ 06.43 &60.05 $\pm$ 00.23  &76.41 $\pm$ 15.23  \\
MS-GCN~\cite{filtjens2022skeleton}&$\triangle$&95.81 $\pm$ 03.43&95.03 $\pm$ 04.04&86.91 $\pm$ 04.97& 87.01 $\pm$ 04.08 \\
\cellcolor{beaublue} Ours &\cellcolor{beaublue}$\star$&\cellcolor{beaublue}\underline{\textit{97.90 $\pm$ 02.48}}&\cellcolor{beaublue}97.15 $\pm$ 02.22&\cellcolor{beaublue}\underline{\textit{91.19 $\pm$ 03.88}} &\cellcolor{beaublue}\underline{\textit{88.85 $\pm$ 03.92}}  \\
\cellcolor{beaublue} Ours (Fusion) &\cellcolor{beaublue}$\star,\square$&\cellcolor{beaublue}\textbf{98.02 $\pm$ 01.71}&\cellcolor{beaublue}\underline{\textit{97.75 $\pm$ 02.39}}&\cellcolor{beaublue}\textbf{91.55 $\pm$ 03.58}  &\cellcolor{beaublue}\textbf{89.40 $\pm$ 02.62}  \\

\specialrule{1pt}{1pt}{1pt}
\end{tabular}

}

\caption{Quantitative comparisons on Desktop Assembly. Note that $\square$ denotes direct RGB inputs, $\triangle$ denotes 3D pose coordinate inputs, and $\star$ denotes 2D pose heatmap inputs. Best results are in \textbf{bold}. Second best results are \underline{\textit{underlined}}.}
\label{tab:comparison-da}
\end{minipage}

\end{table*}

\begin{figure*}[t]
    \centering
	\includegraphics[width=0.55\linewidth, trim = 0mm 100mm 105mm 0mm, clip]{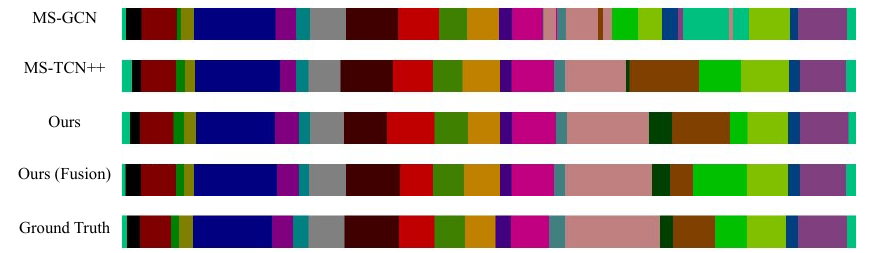}
    \caption{Qualitative comparisons on Desktop Assembly (sequence \emph{2020-04-02-150120}).}
    \label{fig:da}
\end{figure*}

We now evaluate our approaches on Desktop Assembly. Tab.~\ref{tab:comparison-da} shows the quantitative results. It can be seen that our multi-modality fusion approach achieves the best overall performance, followed by our 2D skeleton-based approach, whereas STL~\cite{parsa2020spatio} performs the worst, e.g., on mAP, $91.55\%$ for Ours (Fusion) and $91.19\%$ for Ours, as compared to $86.91\%$, $60.05\%$, and $91.00\%$ for MS-GCN~\cite{filtjens2022skeleton}, STL~\cite{parsa2021multi}, and MS-TCN++~\cite{li2020ms} respectively. The above results show the advantages of using 2D skeleton heatmaps and TCNs for learning spatiotemporal features. Lastly, we present some qualitative results in Fig.~\ref{fig:da}, where our results match the ground truth better than MS-GCN~\cite{filtjens2022skeleton} and MS-TCN++~\cite{li2020ms}. Please see also our supplementary video~\footnote{Supplementary video: \url{https://youtu.be/skx7rkkhcUw}}.

\subsection{Discussions}
\label{sec:exp_discussions}

\noindent \textbf{Numbers of Parameters.} 
We discuss the numbers of parameters of our approaches and previous methods on UW-IOM. Our heatmap-only model with around 1M parameters is larger than MS-GCN~\cite{filtjens2022skeleton} with around 650K parameters, but significantly smaller than STL and MTL~\cite{parsa2021multi} with about 20M and 40M parameters respectively. MS-TCN++~\cite{li2020ms} has the same number of parameters (i.e., about 1M parameters) as our heatmap-only model, while the number of parameters of our fusion model roughly doubles that of our heatmap-only model (yielding about 2M parameters).

\noindent \textbf{Run Times.} We measure the run times of our approaches and previous methods on UW-IOM. Our heatmap-only approach has a similar run time as MS-GCN~\cite{filtjens2022skeleton}, i.e., 71ms and 72ms respectively. They are considerably more efficient than STL and MTL~\cite{parsa2021multi}, which have run times of 183ms and 554ms respectively. The run time of our fusion approach (i.e., 147ms) roughly doubles that of our heatmap-only approach.

\noindent \textbf{Limitations.} 
Despite our state-of-the-art performances on standard action segmentation datasets, our 2D skeleton heatmap-based approach may suffer from a few drawbacks. In contrast to 3D skeletons, 2D skeletons do not include depth cues. Thus, our approach may fail in cases where depth cues are important, e.g., occlusions and viewpoint changes. However, depth cues can still be inferred implicitly from 2D skeleton heatmaps, similar to monocular depth prediction. In addition, both 3D and 2D skeletons do not contain context details, e.g., objects and background information, which are available in RGB videos. Therefore, our approach may suffer in scenarios where context details are crucial. Nevertheless, our fusion approach overcomes that by utilizing both 2D skeleton heatmaps and RGB videos as inputs.

\section{Conclusion}
\label{sec:conclusion}

We introduce a 2D skeleton-based action segmentation method, which uses 2D skeleton heatmap inputs and employs TCNs for spatiotemporal feature learning. This is in contrast with previous methods, where 3D skeleton coordinates are handled directly and GCNs are used to capture spatiotemporal features. Our approach achieves similar/better results and higher robustness against missing keypoints than previous methods on action segmentation datasets, without requiring 3D information. To further improve the results, we fuse 2D skeleton heatmaps with RGB videos. To our best knowledge, this work is the first to utilize 2D skeleton heatmap inputs and the first to perform 2D skeleton+RGB fusion for action segmentation. Our future work will study the generalization of our approach by evaluating it on Epic Kitchens~\cite{damen2020epic}, which has diverse hand-object interactions and camera viewpoints.

{\small
\bibliographystyle{IEEEtran}
\bibliography{references}
}

\end{document}